\documentclass[runningheads]{llncs}
\usepackage{amsmath,amssymb,amsfonts}

\let\accentvec\vec

\let\vec\accentvec 
\usepackage{amsmath}
\usepackage{amssymb}

\usepackage{amsthm}
\usepackage{times}
\usepackage{breakurl}
\usepackage{array}
\usepackage{verbatim}
\usepackage{algorithm}
\usepackage{bbding}
\usepackage{algpseudocode}
\usepackage{lettrine}
\usepackage{booktabs}
\usepackage[T1]{fontenc}

\usepackage{mathrsfs}
\usepackage{graphicx}
\usepackage[backref]{hyperref}
\hypersetup{
hidelinks,
colorlinks=true,
linkcolor=black,
citecolor=black
}
\usepackage{color}

\begin{document}
\title{CAF-YOLO: A Robust Framework for Multi-Scale Lesion Detection in Biomedical Imagery}
%
%

\author{Zilin Chen\inst{(}\Envelope\inst{)}, Shengnan Lu}
\authorrunning{F. Author et al.}
%
\institute{School of Computer Science, Xi’an Shiyou University, Xian, China\\
\email{\{chenzilin0925\}@163.com}}
\maketitle              
\begin{abstract}
Object detection is of paramount importance in biomedical image analysis, particularly for lesion identification. While current methodologies are proficient in identifying and pinpointing lesions, they often lack the precision needed to detect minute biomedical entities (e.g., abnormal cells, lung nodules smaller than 3 mm), which are critical in blood and lung pathology. To address this challenge, we propose CAF-YOLO, based on the YOLOv8 architecture, a nimble yet robust method for medical object detection that leverages the strengths of convolutional neural networks (CNNs) and transformers. To overcome the limitation of convolutional kernels, which have a constrained capacity to interact with distant information, we introduce an attention and convolution fusion module (ACFM). This module enhances the modeling of both global and local features, enabling the capture of long-term feature dependencies and spatial autocorrelation. Additionally, to improve the restricted single-scale feature aggregation inherent in feed-forward networks (FFN) within transformer architectures, we design a multi-scale neural network (MSNN). This network improves multi-scale information aggregation by extracting features across diverse scales. Experimental evaluations on widely used datasets, such as BCCD and LUNA16, validate the rationale and efficacy of CAF-YOLO. This methodology excels in detecting and precisely locating diverse and intricate micro-lesions within biomedical imagery. Our codes are available at \url{https://github.com/xiaochen925/CAF-YOLO}.

\end{abstract}
\section{Introduction}
\begin{figure}[t]
    \centering
    \includegraphics[width=1\linewidth,height=0.23\linewidth]{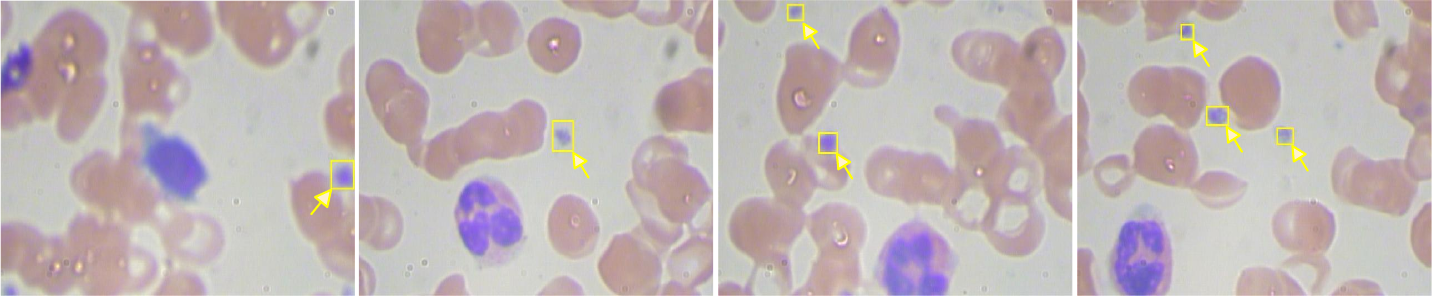}
    \caption{Examples of platelets that are difficult to visually discern in blurry blood microscopic images. 
 }
    \label{introduction}
\end{figure}

In recent years, significant advancements have been made in applying object detection technology to biomedical imaging. As illustrated in Fig. \ref{introduction}, identifying minute platelets (highlighted by the yellow box) in this blurred blood microscopic image is challenging without assistance. Notably, in fields such as blood microscopy and pulmonary CT imaging, object detection algorithms have enabled precise localization and identification of lesions and pathological anomalies, including cancerous cells and pulmonary nodules. This progress highlights the substantial potential of object detection methodologies in enhancing the efficiency of medical diagnostics and treatment. Consequently, leveraging the unique features of CAF-YOLO, specifically designed for the accurate detection of lesions and pathological irregularities, is of paramount importance in the study of various medical conditions. These efforts promise to provide medical practitioners with improved diagnostic precision and expanded therapeutic options for patients suffering from these diseases.

Object detection methods~\cite{r29,r31} in biomedical imaging commonly fall into two broad categories: Transformer-based \cite{r20} and CNN-based \cite{r2} methodologies. Transformer-based approaches for object detection are an emerging field of research that leverages the powerful semantic modeling capabilities and self-attention mechanisms intrinsic to Transformer architecture to enhance the effectiveness of object detection tasks. 
Notably, DETR (DEtection TRansformer) \cite{r16} has made a seminal contribution by fundamentally reshaping the object detection paradigm. DETR reframes object detection as a set prediction challenge, eliminating the need for conventional anchor boxes and Non-Maximum Suppression (NMS) techniques. CFIL \cite{r28} proposes a frequency-domain feature extraction module and feature interaction in the frequency domain to enhance salient features.
MFC \cite{r30} proposes a frequency-domain filtering module to achieve dense target feature enhancement.
Despite these advancements, several challenges remain in Transformer-based object detection methodologies. The single-scale feature aggregation inherent to the feed-forward network (FFN) within Transformer architecture is inherently limited. Some methodologies employ depth-wise convolutional techniques to enhance local feature aggregation within the FFN. 
However, the increased number of channels within the hidden layer poses a significant challenge, limiting the ability of single-scale token aggregation to fully exploit the richness of channel representations.

Convolutional neural networks (CNNs) are widely employed deep learning models~\cite{r22,r21} known for their proficiency in capturing spatial information within images and extracting high-level features. Single-stage methods such as YOLO \cite{r3,r4,r5} and SSD \cite{r6} are successful CNN-based object detection frameworks applied in various biomedical imaging contexts, such as microscopy imaging. YOLO, a series of classic and continually innovative object detection frameworks, is renowned for its fast and high-precision real-time performance. Joseph Sobek et al. \cite{r7} developed Med-YOLO, which rapidly and accurately identifies and labels large structures in 3D biomedical images, circumventing the time-consuming nature of traditional segmentation models.
While convolutional kernels excel at capturing local features, they inherently lack the capacity for long-range information interaction. However, the integration of convolution and attention mechanisms shows promise as a solution to this limitation. Additionally, training object detection models requires a substantial amount of annotated data, which may not always be practical due to considerations such as patient privacy and ethical concerns.

\begin{figure*}[t]
    \centering
    \includegraphics[width=1\textwidth,height=0.25\textwidth]{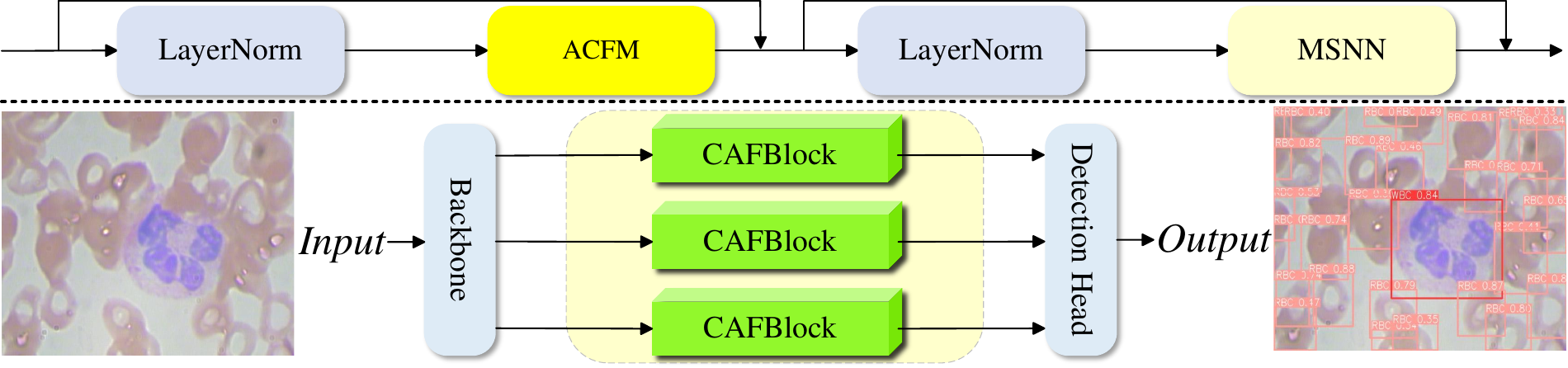}
\caption{Illustration of the overall framework of our proposed CAF-YOLO for biomedical image detection. The internal structure of the CAFBlock is illustrated at the top, comprising two layers of layer normalization, an attention-based convolutional fusion module (ACFM), and a multi-scale neural network (MSNN).
We introduce a CAFBlock after the backbone network to facilitate the extraction of both global and local feature representations.
 }
\label{frame}
\end{figure*}

In response to this challenge, we propose CAF-YOLO, leveraging the YOLOv8 architecture to offer a versatile and robust approach tailored for medical object detection, as illustrated in Fig. \ref{frame}. CAF-YOLO strategically integrates convolutional neural networks (CNNs) and Transformers~\cite{r33}, creating a synergistic fusion of these powerful methodologies. To address the limitation of convolutional kernels' constrained capacity to interact with distant information, we introduce an attention and convolution fusion module (ACFM). This module is meticulously designed to enhance the modeling of both global and local features, thereby facilitating the capture of long-term feature dependencies and spatial autocorrelation.
Furthermore, to overcome the restricted single-scale feature aggregation inherent in feed-forward networks (FFNs) within transformer architectures, we propose a multi-scale neural network (MSNN). This innovative architecture is specifically engineered to enhance multi-scale information aggregation by extracting features across a spectrum of scales, thus addressing the limitations of single-scale aggregation in FFNs. This methodology demonstrates proficiency in accurately detecting and precisely localizing various intricate micro-lesions within biomedical images.

The contributions of this paper can be summarized as follows:

\begin{itemize}
\item We address the complex task of detecting tiny biomedical objects by proposing CAF-YOLO, a method designed to enhance detection capabilities for small biological entities.
\item We introduce CAFBlock, which comprises an attention and convolution fusion module alongside a multi-scale neural network. This innovative architecture adeptly captures global and local features across varying scales while effectively reducing noise.
\item We rigorously evaluate the effectiveness of our proposed approach on two benchmark datasets, BCCD and LUNA16, demonstrating superior performance in detecting tiny biomedical entities with CAF-YOLO. Furthermore, we have made our code publicly available to support further research in the field.
\end{itemize}

\begin{figure*}[t]
    \centering
    \includegraphics[width=1\textwidth,height=0.3\textwidth]{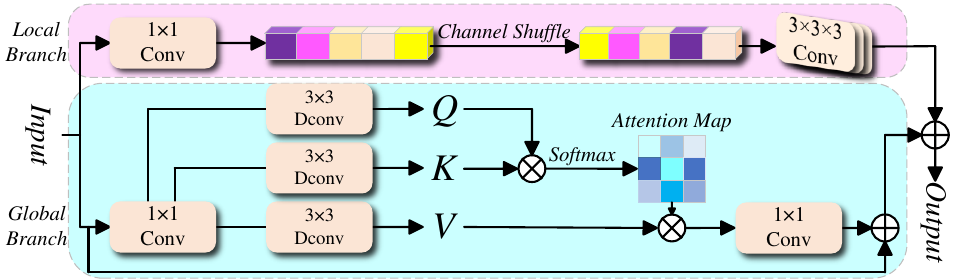}
    \caption{Illustration of the proposed attention and convolution fusion module comprising local and global branches. Within the local branch, convolutional operations and channel shuffling are applied to facilitate the extraction of localized features. Conversely, an attention mechanism is employed within the global branch to model and encapsulate long-range feature dependencies effectively.}
    \label{ACFM}
\end{figure*}
\section{Related Work}
\subsection{Medical Object Detection}  

In recent years, significant advancements have been made in object detection within biomedical images. Many methodologies leverage the domain expertise of medical practitioners to develop specialized feature extractors or utilize extensive biomedical image datasets \cite{r1,r8} for training models aimed at diagnosing lesions and organs. Tasks such as tumor segmentation \cite{r11,r10} and pulmonary nodule detection \cite{r12} hold substantial promise for practical application in biomedical image analysis, aiding clinicians in diagnosing ailments and formulating treatment strategies. However, these approaches are limited by the incompleteness of their manually crafted features and their restricted capacity for broad generalization.

Recently, there has been an increasing focus on object detection within 3D biomedical imagery. Notably, Baumgartner et al. introduced nnDetection \cite{r9}, a framework that automates the configuration process for medical object detection, providing adaptability to diverse medical detection tasks without manual intervention. While 3D biomedical images \cite{r13} contain richer information, their intricate structures impose greater demands on model performance and efficiency. Moreover, many existing 3D detection models \cite{r9,r7} struggle to effectively identify extremely small cancerous cells or rare pathological structures. Consequently, we revisit conventional 2D image detection methodologies and introduce CAF-YOLO as a solution to address detection challenges similar to those encountered in blood microscopy imaging and the identification of pulmonary microlesions.

\subsection{YOLO Series}
YOLOv8, an evolution of YOLOv5, marks a significant advancement in real-time object detection. It combines the foundational strengths of its predecessors with enhancements in network architecture, training procedures, and feature extraction capabilities. YOLOv8 achieves superior precision and efficiency, setting new benchmarks across various datasets.
YOLOv8 operates by dividing the input image into a grid and using a pre-trained convolutional neural network (e.g., Darknet-53 \cite{r14}, ResNet \cite{r15}) for feature extraction. The resulting feature maps are divided into grid cells, each responsible for detecting objects. Predictions include bounding boxes, object classes, and confidence scores, with parameters defining the box's location and size.

Leveraging YOLOv8's detection performance, we integrated CAFBlock into its architecture to create CAF-YOLO. This integration enhances feature extraction by combining global and local features through the attention and convolution fusion module within CAFBlock. Additionally, a multi-scale neural network extracts features across diverse scales, improving multi-scale information aggregation and denoising capabilities.

\begin{figure}[t]
    \centering
    \includegraphics[width=1\linewidth,height=0.3\linewidth]{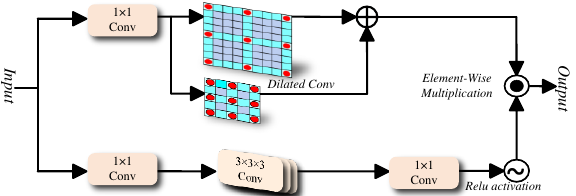}
    \caption{Illustration of the multi-scale neural network (MSNN). In the lower pathway, depthwise convolution is harnessed to facilitate feature extraction, operating with a focus on spatial intricacies. Conversely, the upper pathway utilizes multi-scale dilated convolutions to achieve feature extraction across multiple scales, enabling the capture of diverse contextual information at varying granularities.}
    \vspace{-0.5cm}
    \label{MSNN}
\end{figure}

\section{Proposed Method}
\subsection{The Framework of CAF-YOLO}
The overall structure of the CAF-YOLO model is shown in  Fig. \ref{frame}. Initially, the medical image undergoes preprocessing to align with the requirements of the CAF-YOLO model. Following this, the image undergoes feature extraction and object detection within a unified neural network framework. CAF-YOLO partitions the input image into a grid and assigns bounding boxes to each grid cell, simultaneously predicting object classes and confidence scores. Utilizing multiple anchor boxes notably augments the model's capacity to precisely localize and categorize objects across various scales. Subsequent post-processing steps, including non-maximum suppression, serve to refine the output by eliminating redundant detections and retaining the most precise bounding boxes. Incorporating advancements within the framework of YOLOv8, we propose the integration of a novel component termed CAFBlock. Comprising two fundamental constituents, namely the Attention and Convolution Fusion Module (ACFM) alongside the Multi-Scale Neural Network (MSNN), each CAFBlock is strategically positioned after the YOLOv8 backbone to enhance the modeling of global and local features.
\subsection{Attention and Convolution Fusion Module}  
Considering the convolution operation's limitation regarding its local receptive field, effectively capturing global features may pose challenges. Conversely, Transformers excel at extracting global features and handling long-range dependencies, facilitated by their attention mechanism. Through the fusion of convolution and attention mechanisms, both global and local features can be adeptly modeled. Motivated by this synergy, we introduce the Attention and Convolutional Fusion Module (ACFM), delineated in  Fig. \ref{ACFM}. Within ACFM, we integrate a self-attention mechanism into the global branch to capture a diverse range of global features. Meanwhile, the local branch augments model complexity via channel shuffling, thereby enhancing representational capacity and mitigating the risk of overfitting.

The proposed Convolution and Attention Fusion Module (CAFM) comprises both global and local branches.
In the global branch, an attention mechanism is introduced to enhance long-distance information interaction. 
The initial step involves facilitating the generation of query ($Q$), key ($K$), and value ($V$) tensors through the utilization of $1\times1$ convolution and $3\times3$ depth convolution operations. This process serves to effectuate a transformation upon the input tensor, thereby engendering three distinct tensors distinguished by dimensions denoted as $\hat{H}\times\hat{W}\times\hat{C}$. Herein, the symbols $\hat{H}$ and $\hat{W}$ symbolize the spatial dimensions, while $\hat{C}$ signifies the count of channels inherent within each tensor.  Next, $Q$ is reshaped to $\hat{Q}\in\mathbb{R}^{\hat{H}\hat{W}\times\hat{C}}$, and $K$ is reshaped to $\hat{K}\in\mathbb{R}^{\hat{C}\times\hat{H}\hat{W}}$. To circumvent the computation of an expansive traditional attention map of size $\mathbb{R}^{\hat{H}\hat{W}\times\hat{H}\hat{W}}$and thereby reduce the computational burden, we normalize $\hat{Q}$ and $\hat{K}$ by softmax to calculate the attention map $\mathrm{A}\in\mathbb{R}^{\hat{C}\times \hat{C}}$. The global branch can be formulated as:
\begin{equation}
  \textrm{$\mathit{f}_{\rm att}$} = W_{1\times1}\textrm{Attention}\left(\hat{Q}, \hat{K}, \hat{V}\right) + \mathit{Y},
\end{equation}
 \begin{equation}
  \textrm{Attention}\left(\hat{Q}, \hat{K}, \hat{V}\right) = \hat{V} \textrm{Softmax$\left( \hat{Q} \hat{K}/\alpha \right)$},  
\end{equation}
where $\mathit{f}_{\rm att}$ is an output of the global branch, $W_{1\times1}$ denotes $1\times1$ convolution, $\mathit{Y}$ is the input feature, and  $\alpha$ serves to modulate the magnitude of the matrix multiplication between $\hat{Q}$ and $\hat{K}$ before the softmax operation is applied.

In the local branch, to better interact and integrate features, we first use $1\times1$ convolution to adjust the channel dimensions.  After this, a channel shuffling operation partitions the input tensor along the channel axis into distinct groups, within which depth-wise separable convolution is deployed to induce the shuffling of channels. This step aims to mitigate the risk of model overfitting while enhancing the model's generalization capability and robustness.  Consequent to this step, the resultant output tensors from each group are juxtaposed along the channel axis, thereby engendering a novel output tensor. Following this, we employ a $3\times3\times3$ convolutional operation to extract features. The local branch can be formulated as:
\begin{equation}
\setlength{\abovedisplayskip}{1ex}
    \mathit{f}_{\rm conv} = W_{3\times3\times3}({\rm CS}(W_{1\times1}(\mathit{Y}))),
\setlength{\belowdisplayskip}{1ex}
\end{equation}
where $\mathit{f}_{\rm conv}$ is an output of the local branch, $W_{3\times3\times3}$ denotes $3\times3\times3$ convolution, and CS represents channel shuffle operation.
\subsection{Multi-Scale Neural Network }
The initial feed-forward network (FFN) within the Vision Transformer architecture comprises two linear layers designed for single-scale feature aggregation. Acknowledging the inherent limitation of information encapsulated within the single-scale feature aggregation of FFN, we introduce a Multi-scale Neural Network (MSNN) to bolster nonlinear transformations. After each ACFM module, the output is directed into the MSNN, facilitating the amalgamation of multi-scale features and augmenting nonlinear information transformation. Prior research endeavors(eg. EfficientDet \cite{r21} ), underscore the efficacy of integrating multi-scale information within the domain of object detection tasks.
The detailed structure of MSFN is illustrated in Fig. \ref{MSNN}. The input features are processed through two parallel paths, both paths use $1\times1$ convolution to adjust the channel dimensions. In the lower path, a $3\times3\times3$ depth-wise convolution is used for feature extraction. Following this, nonlinear properties are introduced via the Rectified Linear Unit (ReLU) activation function. In the upper path, to augment the receptive field and facilitate the extraction of a broader spectrum of features, we employ two $3\times3$ dilated convolutional layers with dilation rates $N1$ and $N2$, respectively, in the experimental setup, we designate $N1$ as 2 and $N2$ as 3. Then, a gating mechanism is introduced to augment the non-linear transformation through an element-wise product operation applied to features originating from both pathways. Finally, we use a $1\times1$ convolution kernel to adjust the dimension for the final output. 
Given an input tensor $\mathit{X}\in\mathbb{R}^{\hat{H}\times \hat{W} \times \hat{C}}$,  MSNN is formulated as:

\begin{equation}
\begin{split}
\textrm{Gating}(\mathit{X}) = {\phi}(W_{3\times3\times3} W_{1\times1}(\mathit{X}))\odot (W^{N1}_{3\times3} W_{1\times1}(\mathit{X})+W^{N2}_{3\times3} W_{1\times1}(\mathit{X})),
\end{split}
\end{equation}

\begin{equation}
\textrm{$\mathit{X}_{out}$} = W_{1\times1}\textrm{Gating}(\mathit{X}),
\end{equation}
where $\phi$ represents the Relu activation, $\odot$ denotes element-wise multiplication, $W^{N1}_{3\times3}$ denotes $3\times3$ dilated convolution with dilation rate of N1 and  $W^{N2}_{3\times3}$ denotes $3\times3$ dilated convolution with dilation rate of N2. The dilation rate should be determined according to the specific task requirements.
\section{Experiment and Analysis}\label{sec:exp}  
To validate the proposed CAF-YOLO method's superiority, it is compared with multiple state-of-the-art object detection approaches on two large-scale datasets, namely, BCCD and Luna16.
\subsection{Datasets}
\textbf{\emph{BCCD}}    (Blood Cell Count and Detection) dataset is a comprehensive, lightweight image collection containing 12,500 real high-resolution microscopy blood sample images. This dataset provides a diverse range of colored blood sample images with a variety of blood cells, including normal and abnormal cells, and captures a variety of shapes, sizes, and staining characteristics. The images represent the four main types of blood cells: red blood cells (RBCs), white blood cells (WBCs), platelets, and combinations of these cells, covering varying fields of view and cell densities. Each image has corresponding annotation information, including the position coordinates and category labels of blood cells. This annotation information can be used for supervised learning and algorithm evaluation.

\textbf{\emph{Luna16}} is the most representative and authoritative high-quality pulmonary nodule CT image data set in current pulmonary nodule detection. This dataset has a total of 888 3D lung CT images, 1186 lung nodules, and 36378 pieces of information annotated by 4 professional radiologists. The data set consists of four parts: original CT images, pulmonary nodule location annotation files, original CT lung region segmentation files, and diagnosis result files. This paper divides the training set, test set, and validation set according to the ratio of 7:2:1.
\subsection{Evaluation Metrics}
The evaluation metrics have several components: precision, recall, and mAP(mean average precision). Precision represents the ratio of the number of samples correctly labeled as positive by the classifier to the number of all samples labeled as positive by the classifier. The recall represents the ratio of the number of samples correctly labeled as positive by the classifier to the number of all true positive samples. mAP is the mean of the AP (Average Precision) across all categories. AP is the area under the Precision-Recall curve, which is used to measure the average Precision under different Recalls.
\subsection{Implementation Details}
Implementing CAF-YOLO leveraged PyTorch 2.11, with standardized input image sizes of 640×480 for both the BCCD and LUNA16 datasets. Network optimization was achieved through Stochastic Gradient Descent (SGD), incorporating a learning rate of 0.01. All our computational experiments were executed on a Linux server featuring a 15 vCPU Intel(R) Xeon(R) Platinum 8474C  and an NVIDIA GeForce RTX 4090D 24GB GPU. A batch size of 8 was utilized for training over 300 epochs, with an early stopping mechanism implemented to mitigate overfitting.

\subsection{Comparison with State-of-the-art Methods} 
\begin{table*}[htbp]
\centering
\caption{Performances of different methods on BCCD Dataset\label{tab1}}
\begin{tabular}{|l|c|c|c|c|} \hline 
  
  \textbf{Methods} & \textbf{mAP@50} & \textbf{mAP@50-95} & \textbf{Recall} & \textbf{Precision} \\ \hline 
  
  RT-DETR \cite{r23}& 0.873 & 0.628 & 0.851 & 0.808 \\ \hline  
  YOLOv5& 0.898 & 0.615 & 0.896 & 0.817 \\ \hline  
  YOLOv7 \cite{r24}& 0.907 & 0.615 & 0.890 & 0.845 \\ \hline  
  YOLOv8 & 0.888 & 0.602 & 0.903& 0.819 \\ \hline  
 YOLOv9 \cite{r25}& 0.910& 0.626& 0.900&0.854\\ \hline  
 ADA-YOLO \cite{r32} & 0.912& 0.630& 0.855 & 0.860\\ \hline 
  
   OURS& \textbf{0.922}& \textbf{0.641}& \textbf{0.912}& \textbf{0.887}\\ \hline
  
\end{tabular}
\end{table*}
\subsubsection{Comparisons on BCCD Dataset}
We compare the proposed method with state-of-the-art object detection method on BCCD dataset, and the results are shown in Tab. \ref{tab1}. We can see that our method outperforms all listed pure object detection methods on the evaluation metrics. Specifically, our method outperforms ADA-YOLO \cite{r32} by 1 \%, 1.1 \% and 2.7 \% in mAP@50, mAP@50-95 and precision, respectively; and outperforms YOLOv5 by 0.6 \% in Recall. This demonstrates that the integration of the CAFBlock into the YOLOv8 architecture, enabling the model to effectively capture both global and local features, results in a substantial performance improvement. 
\begin{table*}[htbp]
\centering
\caption{Performances of different methods on LUNA16 Dataset\label{tab2}}
\begin{tabular}{|l|c|c|c|c|} \hline 
  
  \textbf{Methods} & \textbf{mAP@50} & \textbf{mAP@50-95} & \textbf{Recall} & \textbf{Precision} \\ \hline 
  
  Faster R-CNN \cite{r26}& 0.824& 0.607& 0.824& 0.817\\ \hline  
  YOLOv5& 0.802& 0.610& 0.747& 0.742\\ \hline  
  YOLOv7 \cite{r24}& 0.812& 0.609& 0.747& 0.826\\ \hline  
  YOLOv8 & 0.834& 0.616& 0.797& 0.837\\ \hline  
 YOLOv9 \cite{r25}& 0.890& 0.614& 0.816&0.840\\ \hline 
  
  OURS& \textbf{0.907}& \textbf{0.617}& \textbf{0.863}& \textbf{0.869}\\ \hline
  
\end{tabular}
\end{table*}
\subsubsection{Comparisons on LUNA16 Dataset}
To assess the generalizability of our proposed method to medical image object detection,  we compare our method with state-of-the-art CNN-based object detection methods on the LUNA16 dataset, and the results are shown in Tab. \ref{tab2}. Specifically, our method performs baseline outperforms YOLOv9 \cite{r25} by 1.7 \%, 2.9 \% in mAP@50 and precision. This can be attributed to the CAFBlock’s ability to effectively combine the strengths of transformers and CNNs, further highlighting the generalizability of our method across medical image detection tasks.
\begin{table*}[htbp]
\centering
\caption{Ablation Analysis of Our Baseline Gradually Including the Newly Proposed Components on BCCD Dataset\label{tab3}}
\begin{tabular}{|l|l|c|c|c|c|} \hline 
  
   ID&\textbf{Component Settings} & \textbf{mAP@50} & \textbf{mAP@50-95} & \textbf{Recall} & \textbf{Precision} \\ \hline 
   
   1&Baseline& 0.888 & 0.602 & 0.903& 0.819 \\ \hline  
  2&+ACFM & 0.913& 0.624& 0.875&0.864\\ \hline  
  3&+MSNN & 0.899& 0.620& 0.861& 0.850\\ \hline  
 4& +GB+MSNN& 0.901& 0.623& 0.872&0.857\\ \hline  
 5& +LB+MSNN& 0.902& 0.629& 0.874&0.843\\ \hline 
  
    6&OURS & \textbf{0.922}& \textbf{0.641}& \textbf{0.912}& \textbf{0.887}\\ \hline
\end{tabular}
\end{table*}
\subsection{Ablation Studies and Analysis}  
The comparison results presented in Tab. \ref{tab1} and Tab. \ref{tab2} , demonstrate that the proposed CAF-YOLO method is superior to many state-of-the-art object detection methods.
In what follows,  the proposed CAF-YOLO method is comprehensively analyzed from different aspects to investigate the logic behind its superiority, as summarized in Tab. \ref{tab3}. 

(1) Effectiveness of ACFM (i.e. ,ID:2).
(2) Influence of MSNN (i.e. ,ID:3).
(3) Impact of Global Branch (i.e. ,GB,ID:4 )in ACFM.
(4)  Impact of Local Branch in ACFM (i.e. ,LB,ID:5). Note that we maintain the same hyperparameters mentioned in Section Implementation Details during the re-training process for each ablation variant.

\textbf{\emph{Effectiveness of ACFM}}. As shown in Tab. \ref{tab3}, we explore the influence of the Attention and Convolutional Fusion Module (ACFM) in our CAF-YOLO. To verify its necessity, we retrain our network without the ACFM (i.e. ,ID:1) and find that, compared with ID:2, increasing the mAP@50 score from 0.888 to 0.913. This demonstrates that integrating the ACFM into the YOLOv8 architecture, enabling the model to capture global and local features effectively, results in a substantial performance improvement.

\textbf{\emph{Influence of MSNN}}. To explore its influence, we retrain our network without the MSNN (i.e. , ID:1) and find that, compared with ID:3, increasing the mAP@50 score from 0.888 to 0.899. This is attributed to our proposal for the design of a multi-scale neural network (MSNN) aimed at improving multi-scale information aggregation through the extraction of features across diverse scales.

\textbf{\emph{Impact of Global Branch and Local Branch in ACFM}}.  As shown in Tab. \ref{tab3}, to explore the impact of the global branch and local branch in ACFM, we compared the ID:1,3 with ID:4,5,  increasing the mAP@50 score from 0.888 and 0.899 to 0.901 and 0.902.  This demonstrates that integrating a self-attention mechanism into the global branch can capture a diverse range of global features and the local branch augments model complexity via channel shuffling, thereby enhancing representational capacity and mitigating the risk of overfitting.
\begin{figure*}[ht]
    \centering
    \includegraphics[width=1\textwidth,height=0.6\textwidth]{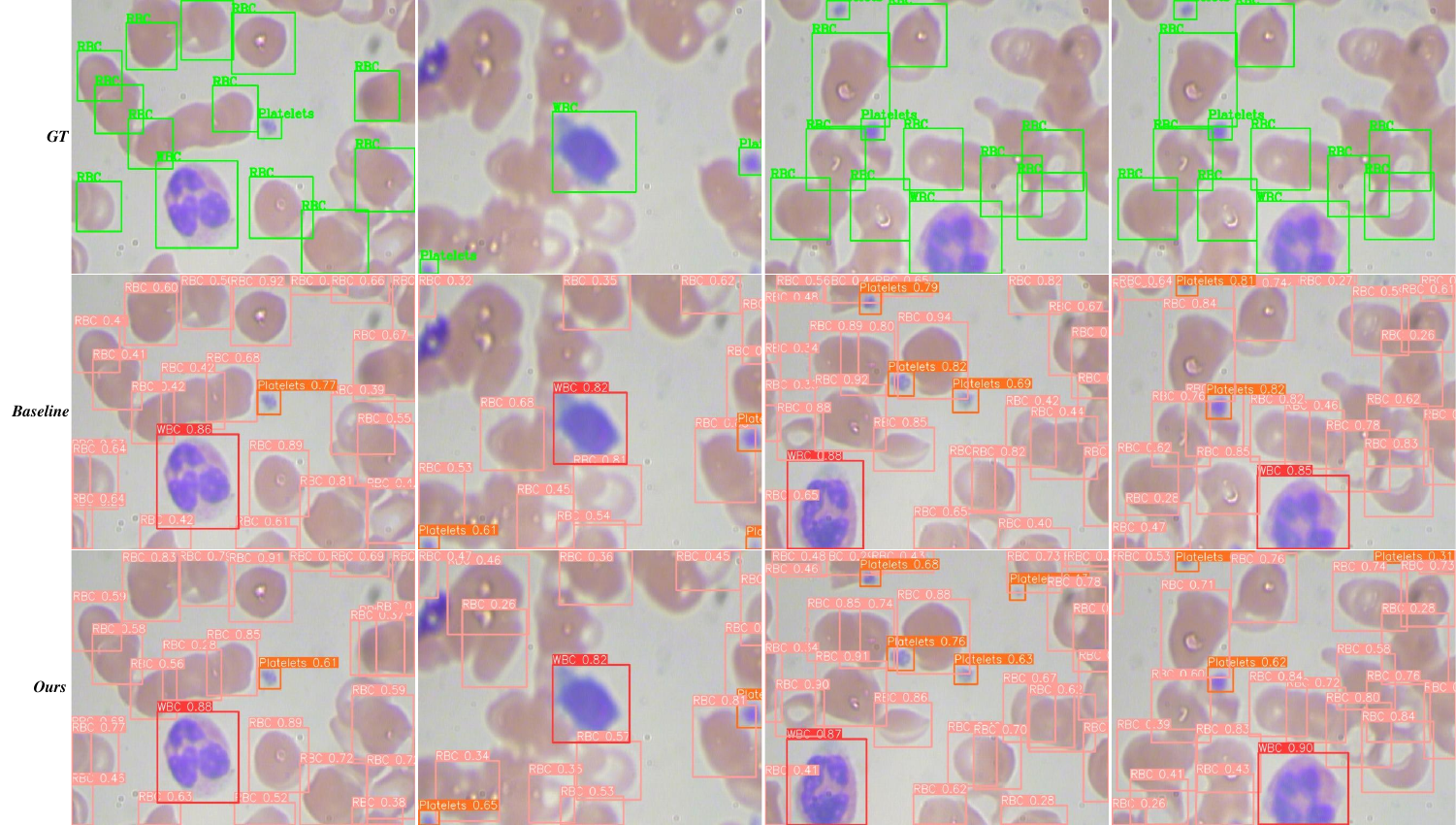}
\caption{The validation set performance  comparison of CAF-YOLO on BCCD dataset can be directly inferred that CAF-YOLO can
successfully detect positive instances of all classes with satisfactory coverage. The image highlights erythrocytes (pink boxes), leukocytes (red boxes), and platelets (orange boxes) with ground truth annotations in green boxes. }
\label{vis}
\end{figure*}
\subsection{Visualization} 
As shown in Fig. \ref{vis}, We compared the image detection results of our method and baseline on the BCCD dataset with the groundtruth. The results show that our method, CAF-YOLO, incorporating an Attention-based Convolutional Fusion Module (ACFM) and a Multi-Scale Neural Network (MSNN), demonstrated superior performance in addressing the challenges of object occlusion and truncation prevalent in medical images. Compared to baseline models, CAF-YOLO successfully detected red blood cells of varying sizes, including minute platelets, achieving comprehensive coverage by effectively identifying positive instances across all classes. This highlights CAF-YOLO’s ability to detect a wider range of positive instances in medical object detection. Our method significantly improved disease detection capabilities, reduced diagnostic errors, and accurately identified a higher proportion of true positive cases, even in the presence of object occlusion and image blur. These findings suggest the potential of CAF-YOLO for earlier and more accurate diagnoses in medical imaging.
\section{Conclusion}\label{sec:con}Object detection plays a crucial role in biomedical image analysis, especially for lesion identification. Despite the proficiency of current methodologies in identifying and pinpointing lesions, they often fall short in detecting minute biomedical entities, such as abnormal cells and lung nodules smaller than 3 mm, which are vital in blood and lung pathology. To address this shortcoming, we have developed CAF-YOLO, based on the YOLOv8 architecture. This method is both nimble and robust, leveraging the strengths of convolutional neural networks (CNNs) and transformers.
To overcome the inherent limitations of convolutional kernels, which struggle with long-range information interaction, we introduced the attention and convolution fusion module (ACFM). This module enhances the modeling of global and local features, enabling the capture of long-term feature dependencies and spatial autocorrelation. Additionally, we designed a multi-scale neural network (MSNN) to improve the limited single-scale feature aggregation found in feed-forward networks (FFN) within transformer architectures. The MSNN enhances multi-scale information aggregation by extracting features across diverse scales.
Experimental evaluations on widely used datasets, such as BCCD and LUNA16, validate the effectiveness and rationale behind CAF-YOLO. The results demonstrate its superior ability to detect and precisely locate various intricate micro-lesions within biomedical imagery. This advancement holds significant potential for improving medical diagnostics and treatment strategies. Our code is publicly available to support further research and development in this field at \url{https://github.com/xiaochen925/CAF-YOLO}.
\bibliographystyle{splncs04}
\bibliography{ref}
\end{document}